%% file: iclr2026_conference.tex
\documentclass{article} % For LaTeX2e
\usepackage{iclr2026_conference,times}

\input{math_commands.tex}

\usepackage{hyperref}
\usepackage{cleveref}
\usepackage{url}

\usepackage{booktabs} % for professional tables
\usepackage{makecell}
\usepackage{colortbl}  
\usepackage{threeparttable}
\usepackage{graphicx}
\usepackage{wrapfig}
\usepackage{float}
\usepackage{bbm}
\usepackage{hyperref}
\usepackage{multirow}
 \usepackage{fontawesome} 
\usepackage[ruled,vlined]{algorithm2e}
\usepackage{algorithmic}  
\usepackage{tikz}  

\definecolor{Gray}{gray}{0.92}
\definecolor{Gray1}{gray}{0.95}
\definecolor{Gray2}{gray}{0.88}
\definecolor{ggray}{RGB}{127,127,127}
\definecolor{aliceblue}{rgb}{0.94, 0.97, 1.0}
\newcommand{\type}[1]{\color{gray}{\small{#1}}}

\title{Mixture of Neuron Experts}

\author{
Runxi Cheng$^{1}$\thanks{Work done during Runxi Cheng’s internships at Microsoft} ,
Yuchen Guan$^{1}$,
Yucheng Ding$^{3}$, 
Qingguo Hu$^{4}$, 
Yongxian Wei$^{1}$, \\
\;\textbf{Chun Yuan$^{1\dagger}$, 
Yelong Shen$^{2}$, 
Weizhu Chen$^{2\dagger}$,
Yeyun Gong$^{2}$\thanks{Corresponding authors: {Yeyun Gong}, Yuan Chun, and Weizhu Chen. \faEnvelopeO~: \texttt{yegong@microsoft.com}; \texttt{yuanc@sz.tsinghua.edu.cn}; \texttt{wzchen@microsoft.com}}~
}
\\ 
~$^1$ Tsinghua Shenzhen International
Graduate School, Tsinghua University \quad $^2$ Microsoft \\
~$^3$ Shanghai Jiao Tong University
~$^4$ School of Informatics, Xiamen University\\
\\
}

\iclrfinalcopy
\begin{document}

\maketitle

\input{section/abstract}
\input{section/introduction}

\input{section/relatedwork}

\input{section/method}

\input{section/experiment}

\input{section/conclusion}

\section*{Ethics statement}
This paper presents work whose goal is to advance the field of large language model. There are many potential consequences of our work, none of which we feel must be specifically highlighted here.

\section*{Reproducibility statement}
The details of datasets, model architectures and hyper-parameters are described
in \Cref{exp setup} and \Cref{Experimental datails}.

\bibliography{iclr2026_conference}
\bibliographystyle{iclr2026_conference}

\clearpage
\input{section/appendix}

\end{document}

%% file: math_commands.tex
\usepackage{amsmath,amsfonts,bm}

\def\eqref#1{equation~\ref{#1}}
\def\1{\bm{1}}

\DeclareMathAlphabet{\mathsfit}{\encodingdefault}{\sfdefault}{m}{sl}
\SetMathAlphabet{\mathsfit}{bold}{\encodingdefault}{\sfdefault}{bx}{n}

%% file: section/abstract.tex
\begin{abstract}
In this work, We first explore whether the parameters activated by the MoE layer remain highly sparse at inference. We perform a sparsification study on several representative MoE models. For each expert, we rank parameters by the magnitude of their activations from the gate projection and progressively prune the activated subset. Pruning up to $60\%$ of parameters within that subset causes only negligible task-performance degradation; substantial drops occur only after more than $90\%$ are removed. We further decompose experts into neuron granular MoE and visualize their activation values, finding that most neuron activations are near zero. This observation motivates us to select only high-activation neuron experts during pretraining. Based on this insight, we propose \emph{Mixture of Neuron Experts} (MoNE). MoNE achieve neuron granular expert select by only applying a simple top-$k$ selection within each expert, incurs negligible latency, and requires no additional routing parameters or inter-expert communication. Extensive experiments demonstrate that MoNE matches traditional MoE performance while activating only $50\%$ of the MoE-layer parameters, and it consistently outperforms traditional MoE when compared at equal numbers of activated parameters. These results suggest that MoNE is a practical approach to improving parameter utilization and inference efficiency in MoE-like models.
\end{abstract}

%% file: section/introduction.tex
\section{Introduction}
Large language models (LLMs)~\citep{dai2024deepseekmoe,bai2023qwen,agarwal2025gpt,team2025kimi} based on Mixture-of-Experts approaches have attracted growing interest in both academic research and industry. The fundamental concept of Mixture-of-Experts (MoE) in large language models entails partitioning a large feed-forward network (FFN) into several smaller subnetworks referred to as experts, where only a subset of expert parameters are activated depending on the input. Unlike dense models that activate all parameters uniformly, MoE models achieve greater computational efficiency through sparse activation patterns.

\begin{figure*}[t]
	\centering
	\includegraphics[width=1.0\linewidth]{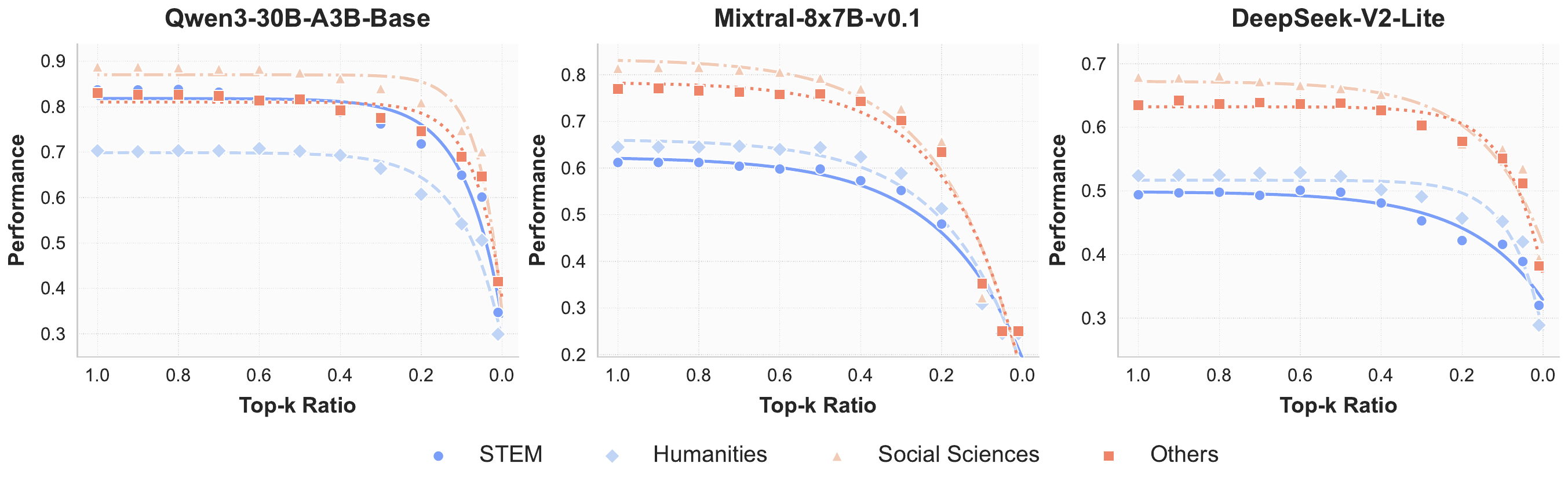}
    \vspace{-12pt}
	\caption{The performance of mainstream MoE models when only use the neuron experts with higher activation weight without extra training. Top-K Ratio refers to the ratio of selected neuron experts.}
        \label{toprate}
        \vspace{-1pt}
\end{figure*}

One key motivation for MoE is the long-observed activation sparsity~\citep{frankle2018lottery,fedus2022sparse1,frantar2023sparse2,frankle2019sparse3} in dense networks: for an given input, only a small fraction of parameters are effectively used. Mixture-of-Experts architectures exploit this property via conditional computation, activating a sparse subset of specialists so as to substantially increase model capacity while maintaining computational efficiency.
This further motivates us to propose the following question:
\begin{center}
    \textit{Are the parameters activated by the MoE layer still highly sparse at inference?}
\end{center}
To answer the question, we performed a sparsification study on a set of representative MoE models. For each expert, we ranked the parameters according to the magnitude of their activations weights, which calculated by the gate projection. Then we progressively pruned the weights from the activated subset according to their rank. The results are presented in \Cref{toprate}. Notably, across three evaluated models, removing up to 60\% of the parameters in this subset led to only negligible declines in task performance, with significant degradation occurring only after more than 90\% were pruned. These results suggest that the parameter subset selected by the MoE gating mechanism still highly sparsity at inference.

\begin{figure*}[t!]
	\centering
	\includegraphics[width=1.0\linewidth]{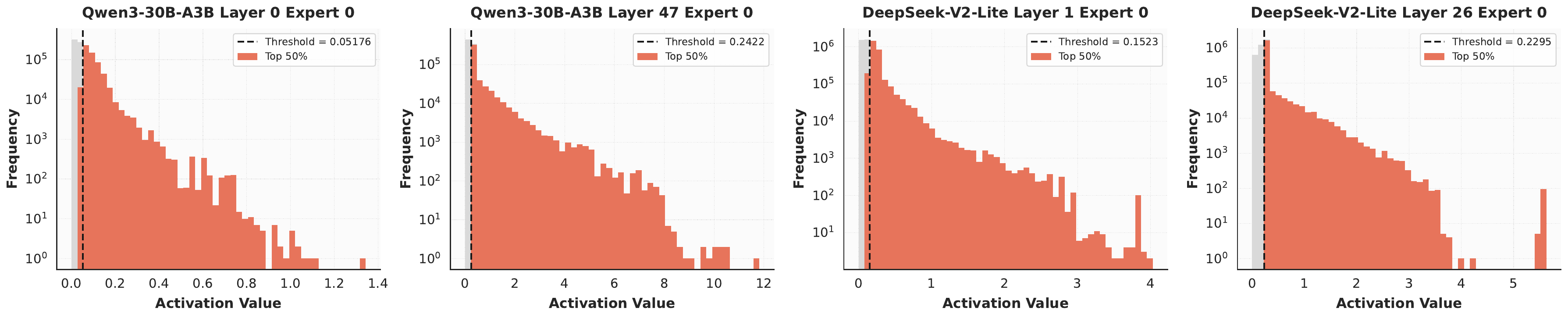}
        \vspace{-12pt}
	\caption{The activation value for the neuron experts, and the top 50\% of these values were highlighted.}
        \label{activation_main}
                \vspace{-10pt}
\end{figure*}

To further explore the sparsity of MoE, we decompose expert into neuron granular MoE. Then we visualize the activation value for the neuron experts. As shown in \Cref{activation_main}, most of the activation values are small, which further demonstrate the sparsity of the MoE layer. This results motivate us to only use the neuron experts with high activation weights for pretraining. Also, recently studies have shown the importance of expert granularity~\citep{krajewski2024scaling,lepikhin2020gshard,du2022glam}: Deepseek V3~\cite{liu2024deepseek} applies 256 experts, Kimi K2~\cite{team2025kimi} applies 384 experts, and Qwen3-Next\cite{qwen3technicalreport} applies 512 experts. However, overly fine-grained expert partitioning requires substantially larger routing networks and incurs significant cross-device communication latency~\citep{lepikhin2020gshard,fedus2022switch}. Therefore, we propose Mixture
of Neuron Experts (MoNE). By applying a simple top-k selection within each expert, we achieve granular selection for MoE without introducing additional router parameters or inter-expert communication. We evaluate MoNE under multiple settings and find that it consistently outperforms traditional MoE.

Our contributions are summarized as follows:

\begin{itemize}
  \item We emprically show that traditionally trained Mixture-of-Experts models exhibit high activation sparsity at inference: a small subset of parameters with large activation values retains most of the model's capability, and most of the activation values are small in the MoE layer.
  \item  We introduce \emph{Mixture of Neuron Experts} (MoNE), which decompose the expert into neuron granular MoE, and achieve neuron granular expert selection via a simple top-k within-expert operation that incurs negligible latency overhead and requires no additional routing parameters
  \item  Extensive experiments demonstrate that MoNE matches the performance of traditional MoE while using only $50\%$ of the parameters in MoE layer. With the same total number of activated parameters, MoNE consistently outperforms traditional MoE.
\end{itemize}

% \begin{figure*}[h]
% 	\centering
% 	\includegraphics[width=1.0\linewidth]{figure/intro_cropped.pdf}
% 	\caption{The input consistency between the pretrained model and the fine-tuned model for ViT-B/32}
%         \label{consistency}
%         \vspace{-5pt}
% \end{figure*}

% \begin{figure*}[tbp]
% 	\centering
% 	\includegraphics[width=1.0\linewidth]{figure/activate.png}
% 	\caption{The input consistency between the pretrained model and the fine-tuned model for ViT-B/32}
%         \label{consistency}
%         \vspace{-5pt}
% \end{figure*}

%% file: section/relatedwork.tex
\section{Related Work}
\subsection{Large Language Models}
Large language models (LLM)~\citep{touvron2023llama,bai2023qwen,gpt3,achiam2023gpt,liu2024deepseek,devlin2019bert,raffel2020t5} have shown remarkable abilities across different tasks, representing important progress toward artificial general intelligence. This success is largely driven by the growth of training data and the expansion of model parameter counts\citep{wei2022emergent,kaplan2020scaling}. And many recent works successfully scaling LLM to billions of parameters~\citep{dai2024deepseekmoe,liu2024deepseek,team2025kimi,agarwal2025gpt,zhang2022opt,scao2022bloom}. However, As model scale increases, the demand for computational resources rises sharply. Consequently, improving the efficiency of both training and inference has become a central research focus to enable further scaling of large language models.

\subsection{Mixture of Experts}
The Mixture of Experts  (MoE)~\citep{cai2025survey,masoudnia2014mixturesurvey,jiang2024mixtral} architecture was introduced to enhance the capacity of deep neural networks while maintaining computational efficiency. \cite{shazeer2017sparsely} proposed integrating an MoE layer between LSTM layers, demonstrating strong performance in language modeling and machine translation tasks. This approach was later adapted into the transformer framework by replacing the standard feed-forward layers with MoE modules. The Switch Transformer~\citep{fedus2022switch} streamlines the expert selection process by assigning each token to only the top-ranked expert, enabling more efficient model scaling. Gshard~\citep{lepikhin2020gshard} refined the routing mechanism by employing a Top-2 expert strategy, leading to substantial improvements in multilingual translation across 100 languages. More recently, DeepseekMoE~\citep{dai2024deepseekmoe} and have introduced fine-grained partitioning of experts within the MoE structure. Grove-MoE~\citep{wu2025grove} incorporating experts of varying sizes. PEER~\citep{he2024mixture} scales the number of experts up to one million. Kimi-K2~\citep{team2025kimi} scales the model to 1,000B parameters and employs 384 experts, while Qwen3-Next~\citep{qwen3technicalreport} further increases the expert count to 512. Improving expert granularity~\citep{tian2025towards,krajewski2024scaling} and increasing the utilization of activated parameters~\citep{li2025slimmoe} are central goals in contemporary MoE research. However, overly fine-grained expert partitioning requires substantially larger routing networks and incurs significant cross-device communication overhead and latency~\citep{lepikhin2020gshard,fedus2022switch}. In this work, we analyze the sparsity of activated parameters in traditional MoE architectures and construct a neuron granularity MoE model. This neuron granularity conversion substantially improves the utilization of activated parameters while avoiding the need for a large router and communication latency associated with overly fine expert partitioning in traditional MoE.

\begin{figure*}[t!]
	\centering
	\includegraphics[width=1.0\linewidth]{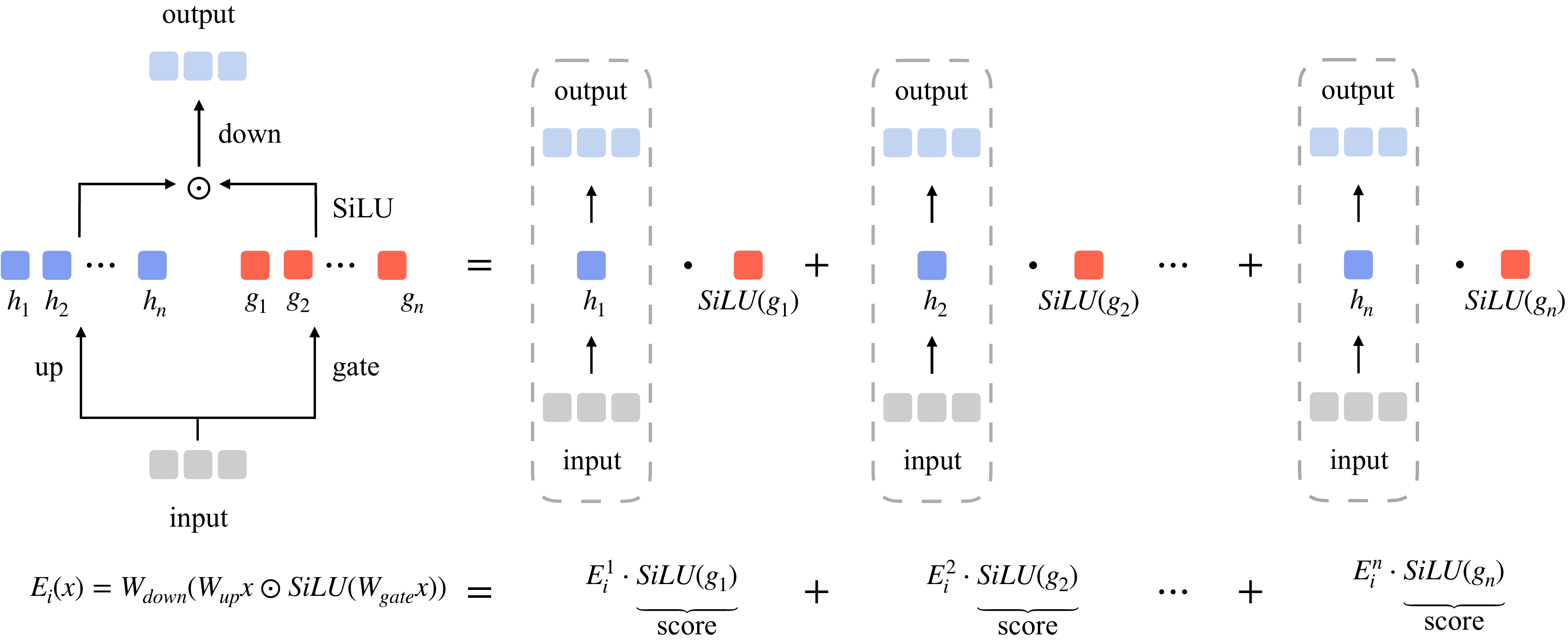}
	\caption{Expert in traditional MoE can be decomposed as the weighted sum of neuron granular FFN, which can be realized as a neuron granular MoE.}
        \vspace{-2pt}
        \label{intro}
    \vspace{-11pt}
\end{figure*}

%% file: section/method.tex
\section{Method}
\subsection{Preliminary on MoE}
The Mixture-of-Experts (MoE) layer extends a standard Transformer by replacing a dense feed-forward network (FFN) with a collection of expert FFNs and a routing mechanism. For each input token, the router conditionally dispatches only a sparse subset of experts, so that computation is performed by a small number of specialists rather than the entire FFN. This sparse execution increases the model’s effective capacity while keeping per-token computation and latency approximately constant, enabling parameter-efficient scaling to much larger models.~\citep{lepikhin2020gshard,dai2024deepseekmoe,fedus2022switch}.

Formally, let $\mathbf{x}\in\mathbb{R}^{d_{\mathrm{model}}}$ denote an input hidden state. An MoE layer consists of $\text{N}_E$ experts $\{E_i\}_{i=1}^{\text{N}_E}$ together with a router that maps $\mathbf{x}$ to routing logits. The
experts typically use the \textit{Gated Linear Unit}~\citep{dauphin2017language} structure, which can be formulated as:
\begin{equation}
    \mathbf{E}_i(\mathbf{x}) = \mathbf{W}_\text{down}^{\,i}(\texttt{SiLU}(\mathbf{W}_\text{gate}^{\,i}\mathbf{x}) \odot \mathbf{W}_\text{up}^{\,i}\mathbf{x})
\end{equation}
where $\mathbf{W}_\text{gate}^{\,i} \in\mathbb{R}^{d_{\mathrm{expert}}\times d_{\mathrm{model}}}$ is the gate projection,
$\mathbf{W}_\text{up}^{\,i} \in\mathbb{R}^{d_{\mathrm{expert}}\times d_{\mathrm{model}}}$ is the up projection, and
$\mathbf{W}_\text{down}^{\,i} \in\mathbb{R}^{d_{\mathrm{model}}\times d_{\mathrm{expert}}}$ is the down projection.
A common routing pipeline is to first calcuate the scores of the router:
\begin{equation}
   \mathbf{P}(\mathbf{x}) \;=\; \texttt{Act}(\texttt{topK}(\text{Router}(\mathbf{x}))), 
\end{equation}
where $\texttt{topK}(\cdot)$ masks out all but the top-K routing logits and $\texttt{Act}(\cdot)$ is the activation function. Then the MoE output is calculated as follows:
\begin{equation}
\mathrm{MoE}(\mathbf{x}) \;=\; \sum_{i=1}^{\text{N}_E}\mathbf{P}(\mathbf{x})_i \, \mathbf{E}_i(\mathbf{x}).
\end{equation}
To encourage balanced utilization across experts, we adopt the commonly used auxiliary load-balance loss:
\begin{equation}
   \mathcal{L}_{\text {aux }}=\alpha_{\text {aux }} \cdot \text{N}_E \cdot \sum_{i=1}^{\text{N}_E} \mathbf{f}_{i} \cdot \mathbf{P}_{i}, \quad\text{where} 
\end{equation}
\begin{equation}
\mathbf{f}_{i}=\dfrac{1}{\text{T}} \sum_{\mathbf{x} \in \mathcal{B}} \mathbbm{1} \{i \in \texttt{argtopK}(\text{Router}(\mathbf{x})) \},\quad
\mathbf{P}_{i}=\dfrac{1}{\text{T}} \sum_{\mathbf{x} \in \mathcal{B}} \texttt{Act}(\texttt{topK}(\text{Router}(\mathbf{x})))[i].
\end{equation}
Here, $\texttt{argtopK}$ get the index of the top-K routing logits, $\mathbf{f}_{i}$ is the fraction of tokens in the batch $\mathcal{B}$ that are assigned to expert $i$, and $\mathbf{P}_{i}$ is the average gating weight that the router assigns to expert $i$ (both estimated over a batch of size $\mathrm{T}$). Minimizing $\mathcal{L}_{\text{aux}}$ therefore penalizes experts that are either under-selected or consistently receive low gating weight, encouraging the router to distribute token assignments and gating weights more evenly across experts. The coefficient $\alpha_{\text{aux}}$ controls the regularization strength, and the multiplicative factor $\mathrm{N}_E$ normalizes the objective with respect to the number of experts.

\begin{algorithm}[tb]
\caption{Mixture of Neuron Experts (MoNE)}
\label{alg:MoNE}
\begin{algorithmic}[1]
\STATE\textbf{Input:}  Layer input $\mathbf{x}\in\mathbb{R}^{d_{\mathrm{model}}}$, number of experts $n$, number of selected experts $\text{K}_E$, number of selected neurons for each expert $\text{K}_N$, activation  function of router $\texttt{Act}$,
\STATE\textbf{Output:} Layer output $\mathbf{h}\in\mathbb{R}^{d_{\mathrm{model}}}$

\STATE $\triangleright$  Initialize $\mathbf{h}=\mathbf{0}$

\STATE $\mathbf{p}= \mathrm{Router}(\mathbf{x}) \in \mathbb{R}^{n}$
\hfill {// Calculate the scores for each expert}

\STATE $\mathbf{I}_E = \texttt{argtopK}(\mathbf{p})$
\hfill {// Select top-$\text{K}_E$ experts}

\STATE $\hat{\mathbf{p}} = \texttt{Act}(\mathbf{p}[\mathbf{I}_E])$
\hfill {// Calculate the activated scores}

\FOR{each selected expert $i\in \mathbf{I}_E$}
  \STATE $\mathbf{G}_i = \texttt{SiLU}(\mathbf{W}_\text{gate}^{i}\mathbf{x})$ \hfill {// Calculate the output of down projection}

  \STATE $\mathbf{I}_N = \texttt{argtopK}(\texttt{Abs}(\mathbf{G}_i))$
  \hfill {// Select top-$\text{K}_N$ neurons}

  \STATE $\tilde{\mathbf{W}}_\text{up}^{\,i} = \mathbf{W}_\text{up}^{i}[\mathbf{I}_N, :]$, $\tilde{\mathbf{W}}_\text{down}^{\,i} = \mathbf{W}_\text{down}^{i}[:, \mathbf{I}_N ]$  \hfill {// Select the weights used for calculation }

  \STATE $\tilde{\mathbf{E}_i}(\mathbf{x}) = \tilde{\mathbf{W}}_\text{down}^{\,i}(\mathbf{G}_i[\mathbf{I}_N] \odot \tilde{\mathbf{W}}_\text{up}^{\,i}\mathbf{x})$ \hfill {// Calculate the output of expert $i$}

  \STATE $\mathbf{h} = \mathbf{h} + \hat{\mathbf{p}}[i]\cdot \tilde{\mathbf{W}}_\text{up}^{\,i}\mathbf{x}$
  \hfill {// Sum the layer output}
\ENDFOR
\STATE \Return $\mathbf{h}$
\end{algorithmic}
\end{algorithm}
\subsection{Mixture of Neuron Experts}
To explore the sparsity within the activated experts, we decompose the each experts into neuron granular mixture of experts. The output of an expert is formulated as:
\begin{equation}
    \mathbf{E}_i(\mathbf{x}) = \mathbf{W}_\text{down}^{\,i}(\texttt{SiLU}(\mathbf{W}_\text{gate}^{\,i}\mathbf{x}) \odot \mathbf{W}_\text{up}^{\,i}\mathbf{x})
    \label{experts}
\end{equation}
For clarity and compactness, we denote the outputs of the gate projection and the up projection by \(\mathbf{G}\) and \(\mathbf{H}\), respectively. Concretely,
\begin{equation}
\mathbf{G} \;=\; \texttt{SiLU}(\mathbf{W}_\text{gate}^{\,i}\mathbf{x})\in\mathbb{R}^{d_{\mathrm{expert}}},\quad
\mathbf{H} \;=\; \mathbf{W}_\text{up}^{\,i}\mathbf{x}\in\mathbb{R}^{d_{\mathrm{expert}}}.
\end{equation}
Then \Cref{experts} can be reformulated as:
\begin{equation}
\mathbf{E}_i(\mathbf{x}) \;=\; \mathbf{W}_\text{down}^{\,i}(\mathbf{G}\odot\mathbf{H}).
\end{equation}
Let $\mathbf{W}_\text{down}^{\,i}[:,k]\in\mathbb{R}^{d_{\text{model}}}$ denote the $k$-th column of $\mathbf{W}_\text{down}^{i}$ and $\mathbf{W}_\text{up}^{i}[k,:]\in\mathbb{R}^{1\times d_{\text{model}}}$ the $k$-th row of $\mathbf{W}_\text{up}$, then we expand the product as follows:
\begin{align}
\mathbf{E}_i(\mathbf{x})
&= \sum_{k=1}^{d_{\text{expert}}}\mathbf{W}_\text{down}^{\,i}[:,k]\,(\mathbf{G}[k]\cdot \mathbf{H}[k]) \\
&= \sum_{k=1}^{d_{\text{expert}}} \mathbf{G}[k]\cdot (\mathbf{W}_\text{down}^{\,i}[:,k]\,(\mathbf{W}_\text{up}^{i}[k,:]\mathbf{x})) \\
&= \sum_{k=1}^{d_{\text{expert}}} \mathbf{G}[k]\cdot (\underbrace{(\mathbf{W}_\text{down}^{\,i}[:,k]\mathbf{W}_\text{up}^{i}[k,:])}_{\;\mathbf{A}_k}\mathbf{x}).
\end{align}
Consequently, we can get the decomposition as follows:
\begin{equation}\label{eq:neuron moe}
\mathbf{E}_i(\mathbf{x}) \;=\; \sum_{k=1}^{d_{\text{expert}}} \mathbf{G}[k]\cdot \mathbf{A}_k\mathbf{x}\;,\quad\text{where}\quad
\mathbf{A}_k \;=\; \mathbf{W}_\text{down}^{\,i}[:,k]\mathbf{W}_\text{up}^{i}[k,:]\in\mathbb{R}^{d_{\text{model}}\times d_{\text{model}}}
\end{equation}
Eq.~\eqref{eq:neuron moe} shows that each expert can be decomposed into a set of neuron granular experts $\mathbf{A}_k$ that weighted by the neuron level activations $\mathbf{G}$. (\textbf{In the subsequent articles, we refer to traditional experts as experts and to neuron granular experts  as neuron experts.}) Motivated by this perspective, we explore the distribution of $\mathbf{G}$ in the mainstream MoE models at inference. As shown in \Cref{activation_main}, the majority of neuron experts receive negligible gate weights: most values of $\mathbf{G}$ are very small, indicating that a large fraction of neurons inside each expert are inactive during inference. To further quantify the impact of these low-activation neurons, we ablate neuron experts whose gate values fall below a threshold and measure the resulting performance. As shown in \Cref{toprate} we find that retaining approximately the top $30\%$ of neuron experts by gate magnitude is sufficient to preserve the bulk of the original performance, which implies that conventionally trained MoE architectures induce high sparsity in the set of activated parameters at inference.

To address this issue, we propose the \emph{Mixture of Neuron Experts} (MoNE). Algorithm \ref{alg:MoNE} formulates the pipeline of MoNE. Concretely, the router first selects a set of experts as usual; for each selected expert $i$, we first calculate the the neuron gating weights $\mathbf{G}$, then we use the weights of neuron experts that associated with high absolute gate values to calcuate the ouput of each experts. MoNE converts the traditional MoE into a neuron granular MoE via a simple single, per-expert sort-and-select operation. In contrast, an explicit neuron level routing design in traditional MoE  would require a substantially larger router and incur heavy cross-expert communication overhead~\citep{lepikhin2020gshard}. MoNE’s selection incurs no additional routing parameters and only accesses the parameters of the expert itself ; because the selected neuron experts communicate within their host expert, the extra communication latency can be negligible. Empirically, pretraining with 50\% of the activation parameters in MoE layer by MoNE already matches the performance of  traditional MoE, which demonstrate MoNE can effectively improve the ultilization of activated parameters.

\subsection{Neuron 
Granular Load Balance Loss}

Since We decompose each expert into neuron level sub-experts, we further introduce the \emph{neuron 
granular load balance loss} (NG-LBL). NG-LBL is designed to avoid cases where a subset of neurons are rarely activated, thereby further improving parameter utilization. The formulation of NG-LBL of is similar with $\mathcal{L}_{\mathrm{aux}}$, but is applied independently to the neuron experts within each expert.

For expert $i$, the fraction of tokens in the batch $\mathcal{B}$ (with batchsize of $\text{T}$) that assigned to neuron $k$, and the average gating weights that the gate projection assigned to neuron $k$ is calculated as follows:
\begin{equation}
\tilde{\mathbf{f}}_{i,k}=\dfrac{1}{\text{T}} \sum_{\mathbf{x} \in \mathcal{B}} \mathbbm{1} \{k \in \texttt{argtopK}(\texttt{Abs}(\mathbf{G}_i)) \},\quad
\tilde{\mathbf{P}}_{i,k}=\dfrac{1}{\text{T}} \sum_{\mathbf{x} \in \mathcal{B}} \texttt{Act}(\texttt{topK}(\mathbf{G}_i))[k].
\end{equation}
The whole auxiliary load balance loss used in MoNE is to sum the orginal auxiliary load balance loss and each expert's neuron granular load balance loss:
\begin{equation}
   \tilde{\mathcal{L}}_{\text {aux }}
= \mathcal{L}_{\text {aux }}
+ \sum_{i=1}^{\text{N}_E} \mathcal{L}_{\text {NG-LBL}}^i,\quad \text{where} \quad 
\mathcal{L}_{\text {NG-LBL}}^i= \alpha_{NG} \cdot d_{\mathrm{expert}} \cdot \sum_{k=1}^{d_{\mathrm{expert}}} \mathbf{f}_{i,k} \cdot \mathbf{P}_{i,k}
\end{equation}
where the $\alpha_{\text{NG-LBL}}$ is the coefficient to controls the regularization strength of NG-LBL.

% Algorithm: Super-Fine-Grained Mixture-of-Experts (with explicit line breaks)

%% file: section/experiment.tex
\input{table/compression}
\input{table/900M}

\section{Experiment}
\subsection{Experimental Setup}
\label{exp setup}

\input{table/rate}

\input{table/act_func}

\textbf{Model Architectures.} As shown in \Cref{tab:models settings}, we implement models with total parameter counts of 925M and 2.81B. The MoE layers for both model scales contained 64 experts. Follow by \cite{dai2024deepseekmoe}, we use one shared expert. The MoE layers in both model scales comprise $64$ experts. For the smaller model, we trained traditional MoE with 4, 6 and 8 experts-per-token, which correspond to activated parameters of 290M, 310M and 330M, respectively; For the larger model, we trained a traditional MoE with 4 experts-per-token, corresponding to 0.55B activated parameters. In the main experimental suite, each traditional MoE configuration was paired with a corresponding MoNE: the $\text{K}_N$ is set to $d_{\mathrm{model}}/4$ and $\text{K}_E$ is set two times of corresponding traditional MoE. so that the number of activated parameters  is equal between the traditional MoE and MoNE. Please refer \Cref{Experimental datails} for detailed training hyper-parameters.

\textbf{Data \& Tokenizer.} We trained the 925M parameter model on a 50B-token subset of the NeMaTron-CC dataset~\citep{su2024nemotron} and the $2.81$B parameter model on a 100B-token subset of the NeMaTron-CC dataset. All text was tokenized with the LLaMA-3-8B tokenizer~\citep{dubey2024llama}. 

\textbf{Hyper-Parameters.} The hyper-parameters are selected based on the common practice for dense language models. We replace all FFN layers with MoE layer in the transformer. Please refer \Cref{Experimental datails} for detailed training hyper-parameters.

\textbf{Benchmarks.} We use the lm-evaluation-harness~\citep{eval-harness} for evaluation. The benchmarks used include ARC-C~\citep{arc-c}, BoolQ~\citep{boolq}, HellaSwag~\citep{hellaswag}, LAMBADA~\citep{lambada}, MNLI~\citep{mnli}, PIQA~\citep{piqa}, RACE~\citep{lai2017race}, SIQA~\citep{siqa}, WinoGrande~\citep{sakaguchi2021winogrande}, WNLI~\citep{levesque2012winograd}. For all these benchmarks, we report the zero-shot accuracy.

\subsection{Main Results}

\textbf{Prior Experiment}
We first compare traditional MoE and MoNE with the same number of experts per token $\text{N}_E$. For MoNE, we set the number of used neurons $\text{K}_{N}$ to $d_{\mathrm{model}}/4$, which reduces the number of MoNE's activated parameters in the moe layer to approximately half of the traditional MoE's. As an additional baseline, we select a random subset of size $K_{N}=d_{\mathrm{model}}/4$ from the neuron experts instead of the $\texttt{TopK}$ strategy to pretrain the model. \Cref{prior_baseline} shows that MoNE attains performance comparable to the traditional MoE while using $50\%$ of the activated parameters in the MoE layer, whereas the random subset baseline suffers a large drop in performance. These findings indicate that MoNE’s selection mechanism effectively improve the utilization of the activated parameters.

\textbf{Comparisons to Traditional MoE of Equivalent Activated Parameters}
We further compared MoNE with Traditional MoE with equivalent activated parameters. As shown in the \Cref{main_result}, MoNE consistently improves over the traditional MoE, and the improvement grows as the number of activated parameters increases. Concretely, when activating 290M parameters MoNE yields an improvement of approximately 1\% relative to the traditional MoE, and this gain increases to about 2\% at 330M activated parameters. Also, with 330M activated parameters, MoNE attains performance that is comparable to a dense model with 925M parameters. For the 3B models MoNE improves upon the traditional MoE by roughly 1.1\%. These results indicate that MoNE is a promising approach for training MoE-like models.

\begin{figure*}[t!]
	\centering
	\includegraphics[width=1.0\linewidth]{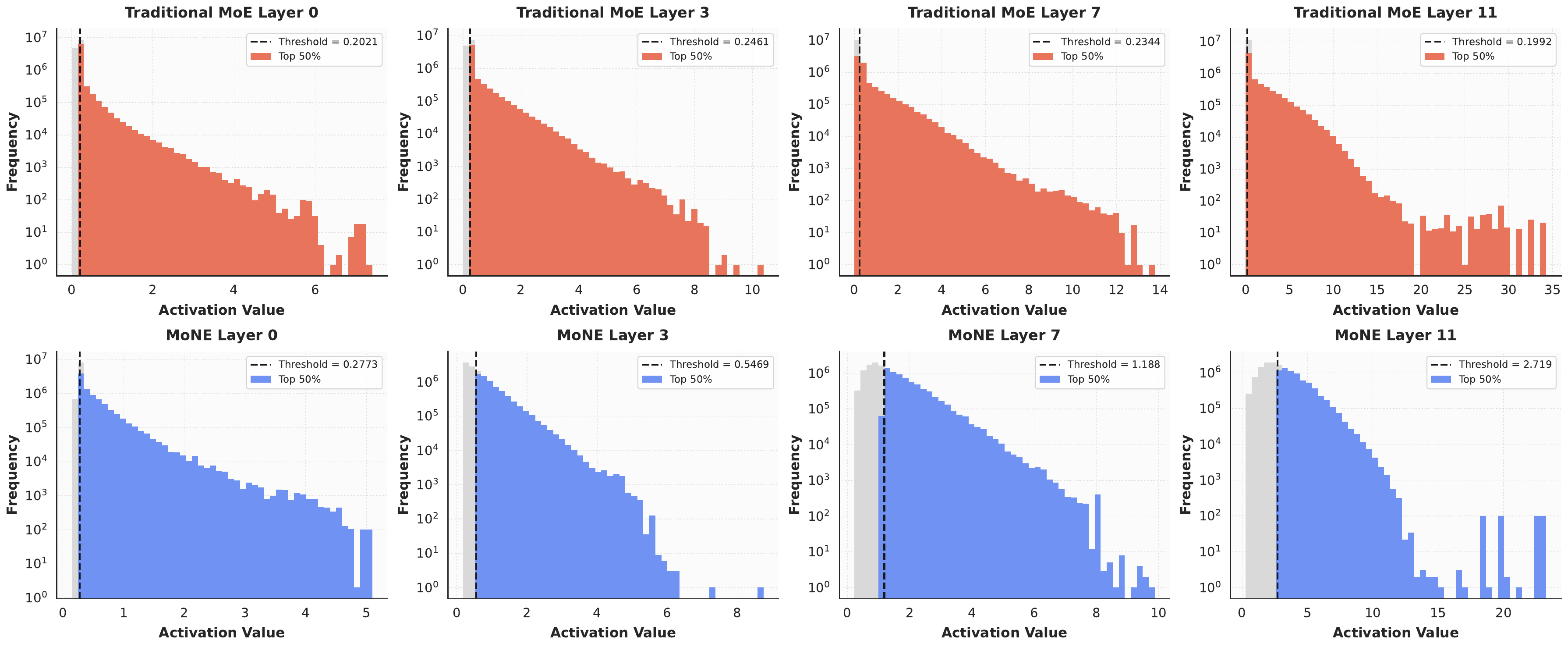}
	\caption{The comparison of the activation value $\mathbf{G}$ for the neuron experts between traditional MoE and MoNE. MoNE effectively increase the activation weight compared with traditional MoE.}
        \label{compare_activation}
    \vspace{-12pt}
\end{figure*}

\subsection{Further analysis}

\begin{figure}[t]
  \centering
  % 左图
  \begin{minipage}[t]{0.35\linewidth}
    \centering
    \includegraphics[width=\linewidth]{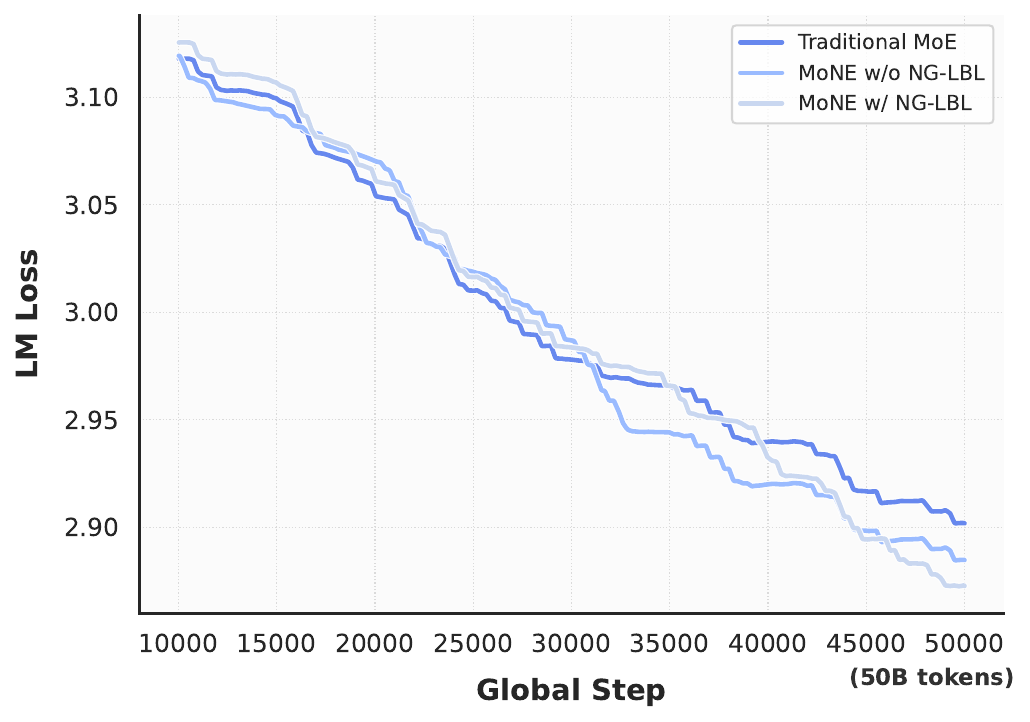}
    \caption{Pre-training loss between traditional MoE and MoNE}
    \label{loss_fig}
  \end{minipage}
  \hfill
  % 右表
  \begin{minipage}[t]{0.63\linewidth}
	\centering
	\includegraphics[width=1.0\linewidth]{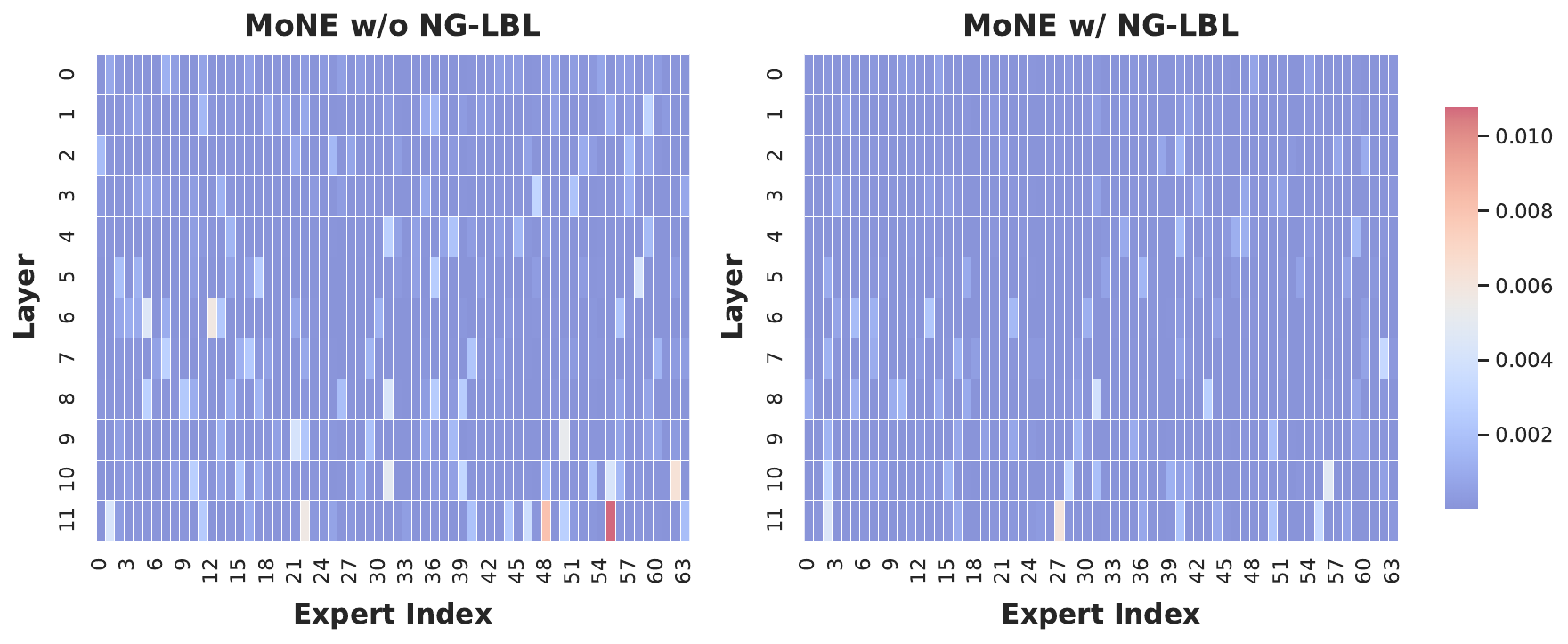}
	\caption{The visualization of load balance for different layers and experts, the value visualized is the variance of $\tilde{\mathbf{f}}_{i,k}$  . }
    \label{lbl}
  \end{minipage}
  \vspace{-12pt}
\end{figure}

\textbf{The Effectiveness of NG-LBL} We investigate the effect of NG-LBL on MoNE in \Cref{main_result} and \Cref{result_ratio}. Empirically, NG-LBL consistently improves MoNE's performance and increases its parameter efficiency: on tue 1B-parameter model, NG-LBL yields an improvement about $1.0\%$, while on a 3B-parameter model the gain is about $1.4\%$. \Cref{loss_fig} shows that NG-LBL substantially accelerates the decline of training loss,which demonstrates the effectiveness of NG-LBL. To better understand how NG-LBL helps, we examine load balancing among neuron experts. As shown in \Cref{lbl}, neuron experts achieve better load balance with  NG-LBL.  This indicate that the balancing at the neuron level can effectively increase the ability of experts

\textbf{The Analysis of the Activation Weight} 
\Cref{compare_activation} compares the activation weight of traditional MoE and MoNE. MoNE effectively increases neuron activation weight: the median  value of activation weight in the first layer rises from 0.20 to 0.28 and continues to grow with depth. The median value increases from 0.20 to 2.70 in the final layer. In addition, the activation weight's distribution produced by MoNE is noticeably more uniform, indicating a more balanced utilization of neuron experts. These results demonstrate that MoNE both increases and homogenizes the use of neuron experts, thereby reducing the sparsity of activated parameters and improving the  utilization of the activated parameters.

\textbf{The Effect of the Neuron Expert  Activation Ratio}
\Cref{toprate} indicates that the within-expert activation rate has a meaningful effect. Keeping the number of activated parameters fixed, we pretrained models with different $\text{K}_N$; \Cref{result_ratio} shows that all settings improve over the baseline, with the best performance at a ratio of $1/4$. Therefore, we recommend using $\text{K}_N = 1/4\cdot d_{\mathrm{model}}$. The result aligns with the result in \Cref{toprate}: model performance remains stable until approximately $70\%$ of neurons are removed, which implies that each activated expert only needs $30\%$ of its neurons to process an input.

\input{table/speed}

\textbf{The Effect of Different Activate function inside the Expert} 
We further explored three internal activation functions for neuron experts in MoNE—\texttt{Sigmoid}, \texttt{SiLU}, and \texttt{Softmax}. Empirically, \texttt{Sigmoid} and \texttt{SiLU} produce consistently performance than \texttt{Softmax}. While experts contain large amounts of neurons, \texttt{Softmax} concentrates probability distribution on a small subset while assigning near-zero weights to most neurons, thereby reducing effective parameter usage. Using NG-LBL mitigates this concentration by encouraging more uniform neuron activation, but \texttt{Softmax} still underperforms the baseline in our experiments. Accordingly, we recommend use \texttt{Sigmoid} or \texttt{SiLU} as the default internal activation for MoNE architectures.

% \begin{figure*}[t!]
%   \centering
%   % 左图
%   \begin{minipage}[t]{0.34\linewidth}
%     \centering
%     \includegraphics[width=\linewidth]{figure/loss.pdf}
%     \caption{Loss}
%     \label{loss_fig}
%   \end{minipage}
%   \hfill
%   % 右图
%   \begin{minipage}[t]{0.64\linewidth}
%     \centering
%     \includegraphics[width=\linewidth]{figure/spf_lbl.pdf}
%     \caption{SPF}
%     \label{load_balance}
%   \end{minipage}

%    % \caption{go}
%   \label{fig:combined}
% \end{figure*}

% \begin{figure*}[t!]
% 	\centering
% 	\includegraphics[width=1.0\linewidth]{figure/spf_lbl.pdf}
% 	\caption{The result on the top rate to select the channel of the gate projection in MMLU}
%         \label{compare_activation}
% \end{figure*}

\textbf{The  Efficiency of MoNE}
We further investigate the efficiency of MoNE. The experiment is conducted on 8 A100s. As shown in \Cref{efficiency}, with the same number of activated parameters, MoNE and traditional MoE require comparable GPU memory and show nearly identical throughput. Crucially, MoNE achieves neuron granular expert selection without enlarging the router or increasing communication latency overhead. Hence, MoNE provides a practical approach to neuron granular expert computation while preserving computation efficiency.

%% file: table/compression.tex
\begin{table*}[t!]
\caption{Comparison between traditonal MoE and MoNE with the same number of activated experts. MoNE shows comparable results while only use half of the activated parameters in the MoE Layer. }
\label{prior_baseline}
\begin{center}
\small
% \begin{sc}
% \cellcolor{blue!7}
\setlength{\tabcolsep}{5pt}
\resizebox{\columnwidth}{!}{
\begin{tabular}{l|cccccccccc|c}
    \toprule[1.25pt]

    \textbf{Model} & \textbf{ARC-C} & \textbf{BOOIQ} & \textbf{HELLA} & \textbf{LAMBDA} &  \textbf{MNLI} & \textbf{PIQA} & \textbf{RACE} & \textbf{SIQA} & \textbf{WINO} & \textbf{WNLI} & \textbf{AVG.} \\

    \midrule

     Traditional MoE \type{4E/64E}& 30.55 & 56.94 & 47.78 & 32.70 & 34.39 & 69.53 & 30.33 & 39.87 & 52.80 & 40.85 & 45.02 \\
     MoNE \texttt{w/} Random Selection  \type{4E/64E}&
27.39 & 56.79 & 37.94 & 27.56 & 35.49 & 65.13 & 30.24 & 38.84 & 49.88 & 43.66 & 42.84 \\
     \rowcolor{Gray} MoNE \texttt{w/} \texttt{TopK} Selection  \type{4E/64E} &
31.31 & 53.64 & 48.08 & 34.76 & 33.96 & 70.02 & 31.48 & 39.25 & 51.78 & 47.89 & \textbf{45.65}

     \\
     \bottomrule[1.25pt]
\end{tabular}
}
% \end{sc}

\end{center}
\end{table*}

%% file: table/900M.tex
\begin{table*}[t!]
\caption{Comparison between traditonal MoE and MoNE with the same number of activated parameters. For 920M parameter and 2.8M LLMs, MoNE exhibits better downstream performance than MoE models.}
\label{main_result}
\begin{center}
\small
% \begin{sc}
% \cellcolor{blue!7}
\setlength{\tabcolsep}{4pt}
\resizebox{\columnwidth}{!}{
\begin{tabular}{l|cccccccccc|c}
    \toprule[1.25pt]

    \textbf{Model} & \textbf{ARC-C} & \textbf{BOOIQ} & \textbf{HELLA} & \textbf{LAMBDA} &  \textbf{MNLI} & \textbf{PIQA} & \textbf{RACE} & \textbf{SIQA} & \textbf{WINO} & \textbf{WNLI} & \textbf{AVG.} \\
    \midrule
    \multicolumn{3}{l}{\emph{925M Activated 925M }} \\
    \midrule
    Dense & 33.28 & 58.62 & 52.07 & 37.05 & 33.49 & 71.27 & 30.72 & 40.99 & 54.22 & 52.11 & 47.84 \\
    \midrule
    \multicolumn{3}{l}{\emph{925M Activated 290M }}\\
    \midrule

     Traditional MoE \type{4E/64E}& 30.55 & 56.94 & 47.78 & 32.70 & 34.39 & 69.53 & 30.33 & 39.87 & 52.80 & 40.85 & 45.02 \\
     \rowcolor{Gray} MoNE \texttt{w/o} NG-LBL  \type{8E/64E} & 30.97 & 55.75 & 48.01 & 33.34 & 34.44 & 70.67 & 29.86 & 38.89 & 53.83 & 49.30 & \textbf{46.01} 
     \\\rowcolor{Gray} MoNE \texttt{w/} NG-LBL \type{8E/64E} & 30.38 & 59.45 & 49.51 & 33.96 & 34.79 & 70.89 & 30.24 & 39.76 & 53.67 & 52.11 & \textbf{47.15}
     \\
     \midrule
     \multicolumn{3}{l}{\emph{925M Activated 310M }}\\
     \midrule
     
     Traditional MoE \type{6E/64E}& {33.02} & {54.50} & {49.20} & {34.48} & {35.04} & {71.33} & {30.91} & {40.58} & {53.43} & {36.62} & 45.12 \\
     \rowcolor{Gray} MoNE \texttt{w/o} NG-LBL  \type{12E/64E}& {30.97} & {55.99} & {50.07} & {35.73} & {32.35} & {69.97} & {30.81} & {40.43} & {51.78} & {52.11} & \textbf{46.58}
     \\\rowcolor{Gray} MoNE \texttt{w/} NG-LBL \type{12E/64E}& {32.17} & {62.11} & {48.65} & {34.58} & {33.89} & {71.44} & {31.00} & {39.20} & {52.88} & {50.70} & \textbf{47.16}
     \\
     \midrule
     \multicolumn{3}{l}{\emph{925M Activated 330M }}\\
     \midrule
     Traditional MoE \type{8E/64E}& {32.68} & {56.61} & {49.70} & {35.51} & {33.36} & {71.93} & {30.33} & {40.74} & {51.22} & {39.44} & {45.43} \\
     \rowcolor{Gray} MoNE \texttt{w/o} NG-LBL \type{16E/64E}& {31.31} & {56.39} & {50.59} & {35.82} & {35.36} & {71.55} & {31.29} & {41.30} & {52.96} & {52.11} & \textbf{47.49} 
     \\\rowcolor{Gray} MoNE \texttt{w/} NG-LBL \type{16E/64E}& {30.38} & {60.31} & {49.17} & {34.58} & {34.82} & {70.84} & {30.81} & {40.94} & {51.38} & {50.70} & \textbf{47.06}
     \\
     \midrule
     \multicolumn{3}{l}{\emph{2.81B Activated 0.55B}}\\
     \midrule
     Traditional MoE \type{4E/64E}& {38.65} & {57.89} & {63.00} & {44.38} & {31.61} & {75.19} & {34.74} & {42.37} & {59.98} & {47.89} & {50.78} \\
     \rowcolor{Gray} MoNE \texttt{w/o} NG-LBL  \type{8E/64E}& {39.93} & {61.13} & {63.87} & {46.26} & {38.67} & {76.12} & {34.70} & {42.82} & {59.19} & {43.66} & \textbf{51.82}
     \\\rowcolor{Gray} MoNE \texttt{w/} NG-LBL \type{8E/64E}& 37.54 & 62.97 & 63.36 & 46.13 & 36.28 & 76.17 & 34.83 & 42.32 & 59.67 & 57.75 & \textbf{53.28}
     \\
     \bottomrule[1.25pt]
\end{tabular}
}
% \end{sc}

\end{center}
\end{table*}

%% file: table/rate.tex
\begin{table*}[t!]
\caption{Architecture exploration on different numbers of selected neurons $\text{K}_N$. MoNE exhibits better downstream performance when the selected rate is $1/4$. }
\tabcolsep=5pt
\label{result_ratio}
\vskip 0.1in
\begin{center}
\begin{small}
% \begin{sc}

\setlength{\tabcolsep}{4pt}
\resizebox{\columnwidth}{!}{
\begin{tabular}{l|cccccccccc|c}
    \toprule[1.25pt]
    \textbf{Model} & \textbf{ARC-C} & \textbf{BOOIQ} & \textbf{HELLA} & \textbf{LAMBDA} &  \textbf{MNLI} & \textbf{PIQA} & \textbf{RACE} & \textbf{SIQA} & \textbf{WINO} & \textbf{WNLI} & \textbf{AVG.} \\

    \midrule
\multicolumn{3}{l}{\emph{2.81B Activated 0.55B}}\\
    \midrule

Traditional MoE \type{4E/64E} & 38.65 & 57.89 & 63.00 & 44.38 & 31.61 & 75.19 & 34.74 & 42.37 & 59.98 & 47.89 & 50.78 \\
    \midrule
\multicolumn{3}{l}{$\text{K}_N = 1/2 \cdot d_{\mathrm{model}} $} \\
    \midrule
    
 MoNE \texttt{w/o} NG-LBL \type{6E/64E }& 41.04 & 57.34 & 63.50 & 46.21 & 36.98  &75.68 & 35.12 & 43.35 & 59.76 & 45.07 & 51.45 \\
\rowcolor{Gray} MoNE \texttt{w/} NG-LBL \type{6E/64E} & 39.59 & 61.96 & 63.10 & 44.89 & 38.41 & 75.30 & 35.22 & 42.17 & 59.12 & 45.07 & \textbf{51.69}\\
    \midrule
\multicolumn{3}{l}{$\text{K}_N = 1/4 \cdot d_{\mathrm{model}} $} \\
    \midrule
MoNE \texttt{w/o} NG-LBL \type{8E/64E}& 39.93 & 61.13 & 63.87 & 46.26 & 38.67 & 76.12 & 34.70 & 42.82 & 59.19 & 43.66 & 51.82 \\
\rowcolor{Gray} MoNE \texttt{w/} NG-LBL \type{8E/64E}& 37.54 & 62.97 & 63.36 & 46.13 & 36.28 & 76.17 & 34.83 & 42.32 & 59.67 & 57.75 & \textbf{53.28} \\
    \midrule
\multicolumn{3}{l}{$\text{K}_N = 1/10 \cdot d_{\mathrm{model}} $} \\
    \midrule
 MoNE \texttt{w/o} NG-LBL \type{10E/64E}& 40.27 & 60.86 & 63.89 & 46.09 & 31.59 & 75.84 & 34.55 & 43.09 & 58.01 & 45.07 & 50.99 \\
\rowcolor{Gray} MoNE \texttt{w/} NG-LBL \type{10E/64E}& 39.85 & 58.10 & 61.99 & 44.21 & 32.07 & 75.03 & 35.12 & 42.12 & 59.27 & 57.75 & \textbf{51.74} \\

    \bottomrule[1.25pt]
\end{tabular}
}
\vspace{-12pt}
% \end{sc}
\end{small}
\end{center}
\end{table*}

%% file: table/act_func.tex
\begin{table*}[t!]
\caption{The performance of MoNE when applying different activation functions. MoNE exhibits
better downstream performance when applying $\texttt{SiLU}$ and $\texttt{Sigmoid}$. }
\tabcolsep=5pt
\label{act_func}
\vskip 0.1in
\begin{center}
\begin{small}
% \begin{sc}
\setlength{\tabcolsep}{4pt}
\resizebox{\columnwidth}{!}{
\begin{tabular}{l|cccccccccc|c}
    \toprule[1.25pt]
    \textbf{Model} & \textbf{ARC-C} & \textbf{BOOIQ} & \textbf{HELLA} & \textbf{LAMBDA} &  \textbf{MNLI} & \textbf{PIQA} & \textbf{RACE} & \textbf{SIQA} & \textbf{WINO} & \textbf{WNLI} & \textbf{AVG.} \\

    \midrule
\multicolumn{3}{l}{\emph{925M Activated 290M}}\\
    \midrule

Traditional MoE \type{4E/64E} & 30.55 & 56.94 & 47.78 & 32.70 & 34.39 & 69.53 & 30.33 & 39.87 & 52.80 & 40.85 & 45.02 \\
    \midrule
\multicolumn{3}{l}{$\texttt{SiLU}$} \\
    \midrule
    
 MoNE \texttt{w/o} NG-LBL \type{8E/64E }& 30.97 & 55.75 & 48.01 & 33.34 & 34.44 & 70.67 & 29.86 & 38.89 & 53.83 & 49.30 & 46.01 \\
\rowcolor{Gray} MoNE \texttt{w/} NG-LBL \type{8E/64E} & 30.38 & 59.45 & 49.51 & 33.96 & 34.79 & 70.89 & 30.24 & 39.76 & 53.67 & 52.11 & \textbf{47.15}\\
    \midrule
\multicolumn{3}{l}{$\texttt{Sigmoid}$} \\
    \midrule
MoNE \texttt{w/o} NG-LBL \type{8E/64E}&  31.40 & 49.45 & 48.93 & 34.76 & 35.00 & 71.33 & 32.73 & 39.51 & 49.57 & 50.70 & 45.78 \\
\rowcolor{Gray} MoNE \texttt{w/} NG-LBL \type{8E/64E}& 30.72 & 59.79 & 47.51 & 33.75 & 34.80 & 70.18 & 32.82 & 40.17 & 51.38 & 53.52 & \textbf{47.10} \\
    \midrule
\multicolumn{3}{l}{$\texttt{Softmax}$} \\
    \midrule
 MoNE \texttt{w/o} NG-LBL \type{8E/64E}& 28.24 & 47.52 & 38.30 & 30.08 & 35.00 & 66.43 & 30.33 & 39.25 & 51.54 & 42.25 & 42.30 \\
\rowcolor{Gray} MoNE \texttt{w/} NG-LBL \type{8E/64E}& 28.58 & 48.53 & 44.96 & 32.52 & 35.16 & 69.91 & 30.43 & 38.89 & 50.59 & 46.48 & \textbf{44.16} \\

    \bottomrule[1.25pt]
\end{tabular}
}
\vspace{-8pt}
% \end{sc}
\end{small}
\end{center}
\end{table*}

%% file: table/speed.tex
\begin{wraptable}{r}{0.48\textwidth}
\centering
\small
\vspace{-8pt}
\caption{Throughput and memory usage comparison among traditional MoE and MoNE. Auxiliary losses do not impact efficiency.}
\label{efficiency}
\vspace{8pt}
\resizebox{0.45\columnwidth}{!}{
\begin{tabular}{lrr}
\toprule[1.25pt]
 & Traditional MoE & MoNE \\
\midrule
\multicolumn{3}{l}{\textbf{Configuration}}\\
Batch size & 8 & 8 \\
Input length & 1024 & 1024 \\
New tokens & 128 & 128 \\
\midrule
\multicolumn{3}{l}{\textbf{Throughput \& Memory}}\\
Tokens/sec & 1340.63 & 1338.49 \\
Memory Peak Reserved& 7.8GB & 7.8GB \\
\bottomrule[1.25pt]
\end{tabular}
}

\end{wraptable}

%% file: section/conclusion.tex
\section{Conclusion}
In this work, we demonstrate that the parameters activated by Mixture-of-Experts (MoE) layers is also highly sparse. By decomposing each expert into neuron granular subexperts.  we find that many neuron experts receive very small activation weights. The result motivate us to improve the utilization of activated paratmeters by only use the neuron experts with high activation weights. Therefore We propose Mixture of Neuron Experts (MoNE), a simple and practical modification of traditional MoE that operates at neuron granularity: by decomposing experts into neuron granular subexperts and applying a simple sorting operation to the gate-projection outputs prior to expert computation, MoNE converts a traditional MoE into a neuron granular MoE. Furthermore, we propose to apply the neuron granular load-balance loss on the neuron experts to encourage more uniform neuron utilization. MoNE requires no additional model parameters and incurs only a negligible computational overhead relative to traditional MoE. Empirically, MoNE matches baseline performance while activating only half of the parameters in the MoE layer and achieves consistent improvements when compared at equal numbers of activated parameters. Expert granularity is an important focus of current MoE development, while traditional MoE faces problems such as large routers and large communication delays when expert partitioning is overly fine granularity,. We believe MoNE is a practical step toward more efficient and scalable MoE-like architectures.

%% file: section/appendix.tex
\appendix
\section{Appendix}
\subsection{Experimental datails}
\label{Experimental datails}

\subsubsection{Model Architectures.}
We list the model configuration in \Cref{tab:models settings}. Here we verified the corresponding MoE and MoNE has the same number of activated parameters. Suppose the number of parameters for \texttt{gate projection}, \texttt{up projection} and \texttt{down projection} is $\text{N}$. For a MoE layer with 4 experts activated, the number of activated parameters is $4 \cdot 3 \cdot  \text{N}$. For a MoNE layer with 6 experts activated and $\text{N}_k$ /$d_\mathrm{model}$ is $1/2$, the number of activated parameters of an expert can be calculated as $\text{N} + 1/2\cdot2\cdot \text{N} $, where the parameters of \texttt{gate projection} are all activated, and the  the parameters of \texttt{up projection} and \texttt{down projection} only activated $1/2$. Then total  activated parameters can be calculated as $6\cdot(\text{N} + 1/2\cdot2\cdot \text{N}) = 12 \text{N}$. Accordingly, for a MoNE layer with 8 experts activated and $\text{N}_k$ /$d_\mathrm{model}$ is $1/4$, the number of activated parameters is $8\cdot(\text{N} + 1/4\cdot2\cdot \text{N}) = 12 \text{N}$, and for a MoNE layer with 10 experts activated and $\text{N}_k$ /$d_\mathrm{model}$ is $1/10$, the number of activated parameters is $10\cdot(\text{N} + 1/10\cdot2\cdot \text{N}) = 12 \text{N}$. Therefore the corresponding MoE and MoNE has the same number of activated parameters.
\input{table/model_arch}

\subsubsection{Hyper-Parameters.}

The hyperparameters  are selected based on the common practice for dense transformer language models~\citep{zhang2024tinyllama,openlm2023openllama,touvron2023llama2,xue2024openmoe}. The key training hyperparameters used in our experiments are as follows: batch size (tokens) $=1\,$M; auxiliary load-balance weight $\alpha_{\text{aux}}=0.001$; neuron-granular load-balance weight $\alpha_{\mathrm{NG}}=0.001$; optimizer = FusedAdam~\citep{kingma2014adam}; learning rate $=5e-4$; router scoring activation function = softmax; weight decay~\citep{loshchilov2017dweightdecay} $=0.1$; model maximum sequence length $=2\mathrm{k}$. These settings were kept fixed across the reported pretraining runs and ablations unless stated otherwise.

\subsubsection{Calculate resources and environment}

We use deepspeed as the training framework. For the 925M model, We conduct training on a cluster with 4 nodes and 32 A100 GPUs. For the 2.81B model, We conduct training on a cluster with 16 nodes and 128 A100 GPUs.

\clearpage
\subsection{Additional Experiment}
\subsubsection{More results on the activation value for the neuron experts.}
\begin{figure*}[h!]
	\centering
	\includegraphics[width=1.0\linewidth]{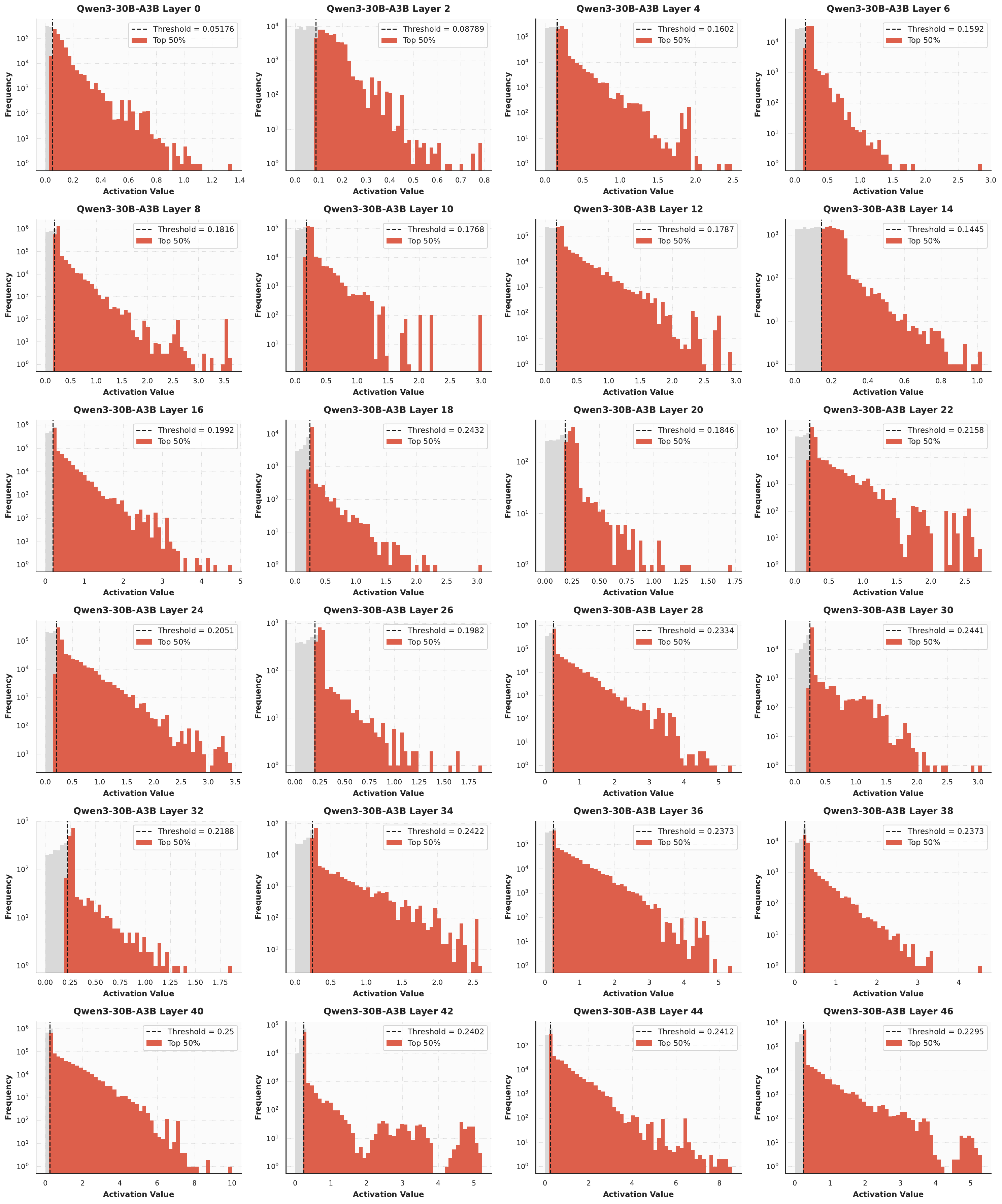}
	\caption{The activation value for the neuron experts on Qwen3-30B-A3}
        \label{more_activation_qwen}
\end{figure*}

\begin{figure*}[h!]
	\centering
	\includegraphics[width=1.0\linewidth]{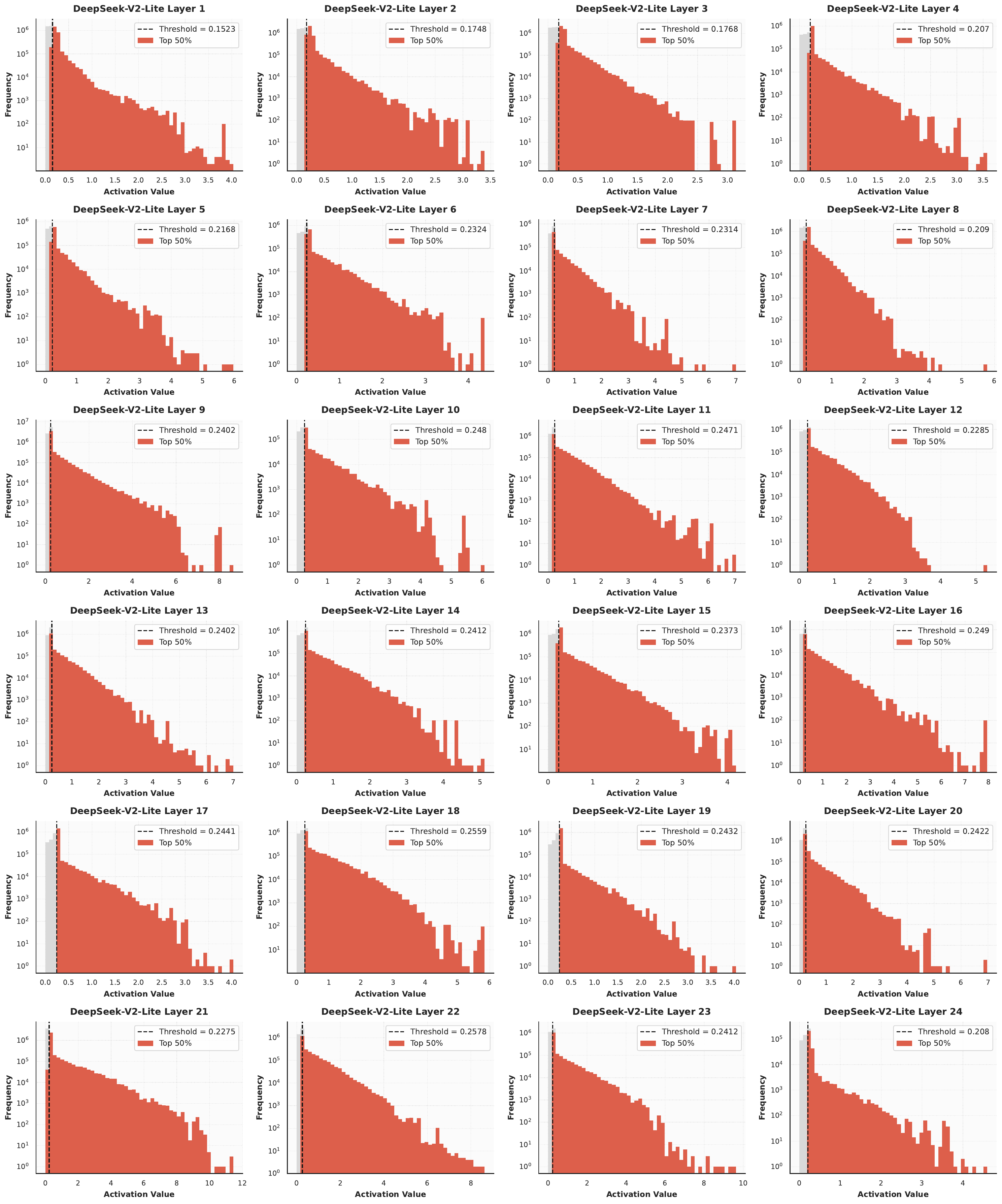}
	\caption{The activation value for the neuron experts on DeepSeek-V2-Lite}
        \label{more_activation_dsv2}

\end{figure*}

We visualize additional neuron granular activation values for Qwen3-30B-A3B and DeepSeek-V2-Lite. As shown in \Cref{more_activation_qwen} and \Cref{more_activation_dsv2}, the vast majority of neuron experts receive negligible gate weights: most entries of $\mathbf{G}$ are close to zero, indicating that a large fraction of neurons within each expert remain effectively inactive during inference.

\begin{figure*}[h!]
	\centering
	\includegraphics[width=1.0\linewidth]{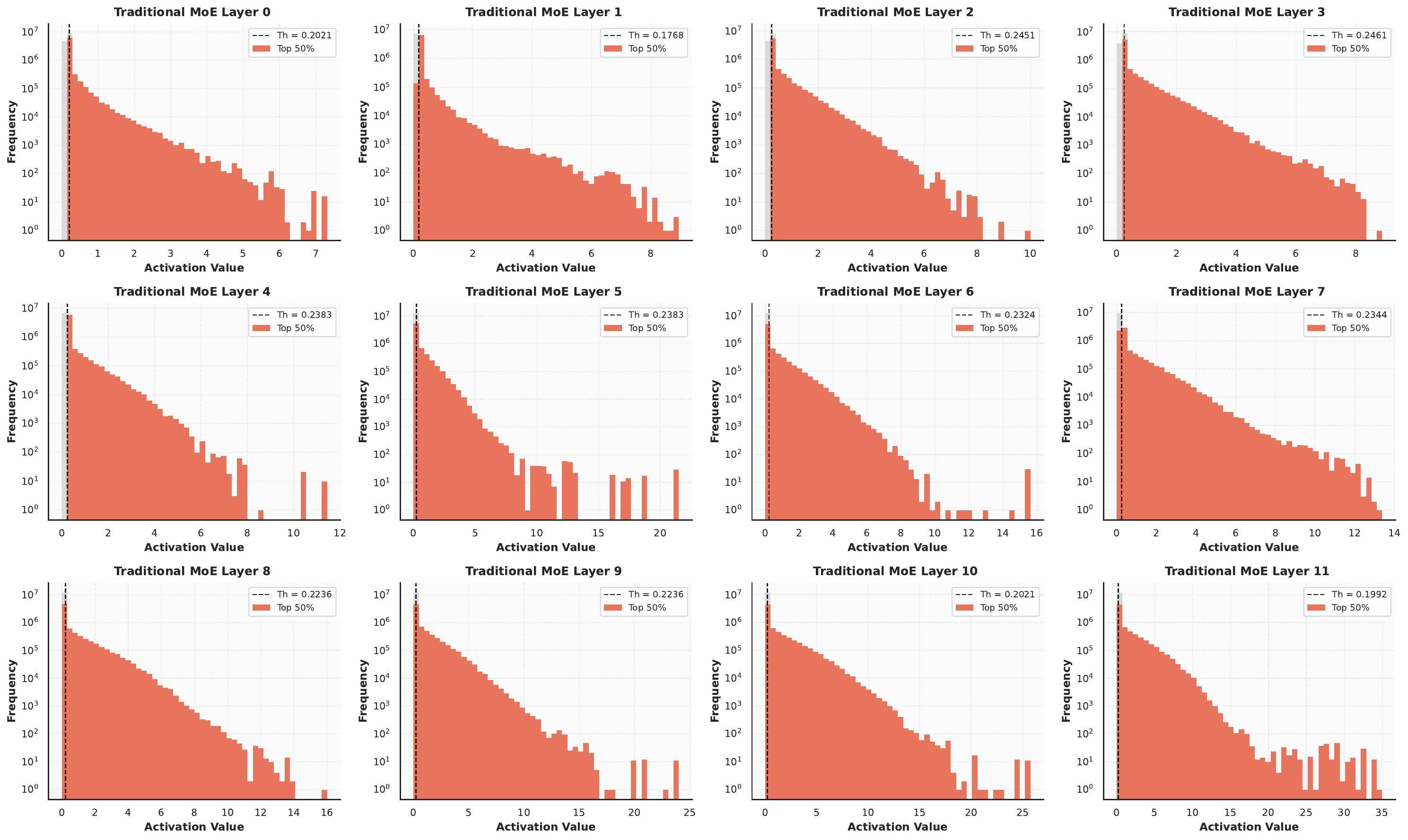}
	\caption{The activation value for the neuron experts on Traditional MoE}
        \label{more_activation_moe}
\end{figure*}

\begin{figure*}[h!]
	\centering
	\includegraphics[width=1.0\linewidth]{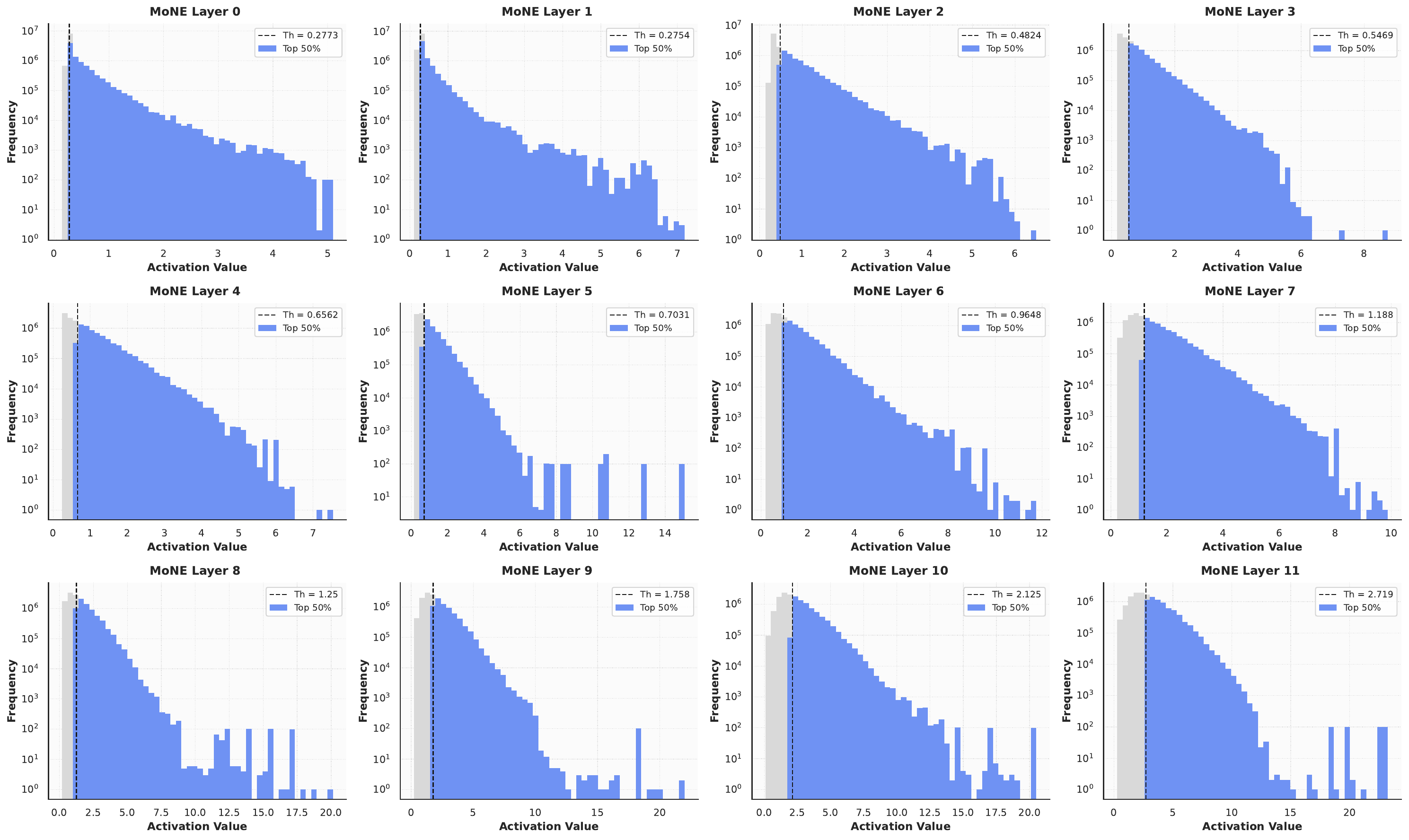}
	\caption{The activation value for the neuron experts on MoNE}
        \label{more_activation_mone}
\end{figure*}

We further compares the activation weight of traditional MoE and MoNE.As shown in \Cref{more_activation_moe} and \Cref{more_activation_mone} MoNE effectively increases neuron activation weight: the median  value of activation weight in the first layer rises from 0.20 to 0.28 and continues to grow with depth. The median value increases from 0.20 to 2.70 in the final layer. In addition, the activation weight's distribution produced by MoNE is noticeably more uniform, indicating a more balanced utilization of neuron experts. These results demonstrate that MoNE both increases and homogenizes the use of neuron experts, thereby reducing the sparsity of activated parameters and improving the  utilization of the activated parameters.

\subsubsection{The comparison of traning loss for traditional MoE and MoNE }

\begin{figure*}[h!]
	\centering
	\includegraphics[width=1.0\linewidth]{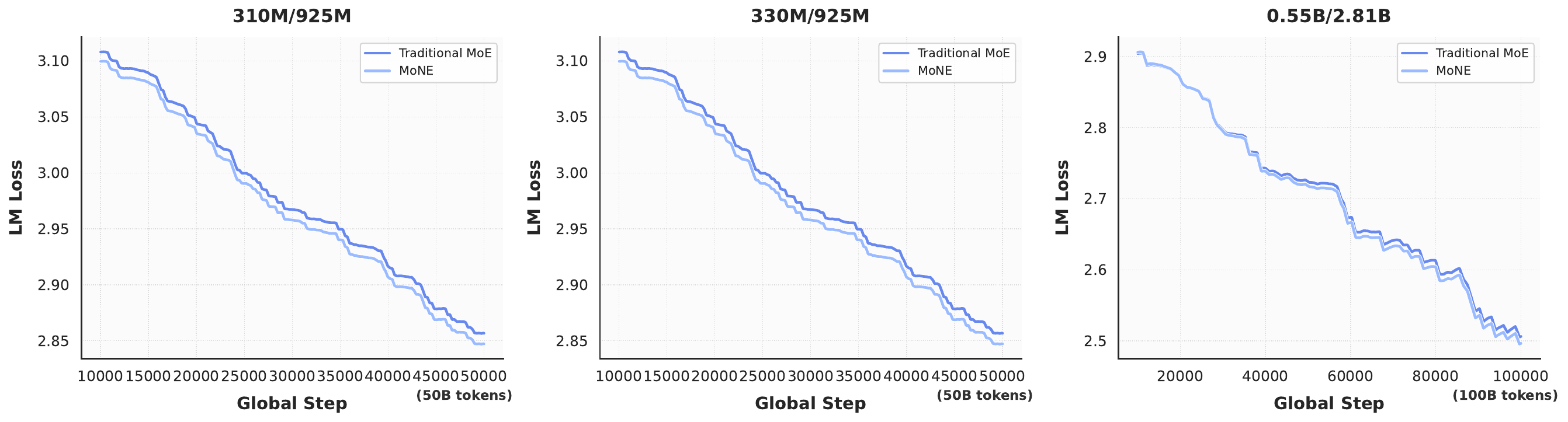}
	\caption{Pre-training loss between
tradition MoE and MoNE.}
        \label{moreloss}
\end{figure*}

As shown in \Cref{moreloss}, MoNE exhibit more effective expert learning compared with traditional MoE, as evidenced by lower
loss values.

\subsubsection{The effect of $\mathcal{L}_{\text{aux}}$ on MoNE}

We further investigate the influence of an auxiliary load-balance loss $\mathcal{L}_{\text{aux}}$ on MoNE. Our experimental results show that $\mathcal{L}_{\text{aux}}$ significantly affects MoNE’s performance, suggesting that the balacnce across experts is important for MoNE to realize further gains in parameter utilization.

\input{section/auxlbl}

\subsection{LLM Usage}
This study utilizes Large Language Models (LLMs) to refine content, adjust formatting, construct
tables, and provide writing suggestions for specific chapters.

%% file: table/model_arch.tex
\begin{table*}[h!]
\scriptsize
\centering
\caption{\small \textbf{Sizes and architectures of MoNE and traditional MoE models.} ``290M/925M'' represents an architecture of an approximately 925M parameter, with 290M activated per token during inference.}
\vspace{8pt}
%\resizebox{1.\linewidth}{!}
\setlength{\tabcolsep}{3.2pt}
{
\resizebox{\columnwidth}{!}{
\begin{tabular}{l|cccccccc}
\toprule[1.25pt]
 \multirow{2}{*}{\textbf{Methods}} & \multirow{2}{*}{{\# Layers}} & {{\# Hidden}} & {{\# Intermediate}} & \multirow{2}{*}{{\# Heads}} & {{\# Head}} & {{\# The Number of}} &  {\# The Number of }  & \multirow{2}{*}{{\# $\text{N}_k$}/$d_\mathrm{model}$} \\ 
  & & {Size} & {Size} &  & { Dim} & {{FFN Experts}} &  {Experts per Token} & \\
 \midrule
 Traditional MoE \type{\scriptsize 290M/925M} & 12 & 768 & 368 & 16  & 48 & 64 & 4 & - \\
 Traditional MoE \type{\scriptsize 310M/925M} & 12 & 768 & 368 & 16  & 48 & 64 & 6  & -\\
 Traditional MoE \type{\scriptsize 330M/925M} & 12 & 768 & 368 & 16  & 48 & 64 & 8 & -\\
 Traditional MoE \type{\scriptsize 0.55B/2.81B} & 24 & 1024 & 512 & 16  & 96 & 64 & 4 & -\\
 MoNE \type{\scriptsize 290M/925M } & 12 & 768 & 368 & 16  & 48 & 64 & 8 & 1/4\\
 MoNE \type{\scriptsize 310M/925M } & 12 & 768 & 368 & 16  & 48 & 64 & 12 & 1/4\\
 MoNE \type{\scriptsize 330M/925M} & 12 & 768 & 368 & 16  & 48 & 64 & 16 & 1/4\\
 MoNE \type{\scriptsize 0.55B/2.81B 6E}  & 24 & 1024 & 512 & 16  & 96 & 64 & 6 & 1/2 \\
 MoNE \type{\scriptsize 0.55B/2.81B 8E}  & 24 & 1024 & 512 & 16  & 96 & 64 & 8 & 1/4 \\
 MoNE \type{\scriptsize 0.55B/2.81B 10E}  & 24 & 1024 & 512 & 16  & 96 & 64 & 10 & 1/10 \\
\bottomrule[1.25pt]
\end{tabular}
}
}
\vspace{-1.2em}
\label{tab:models settings}
\end{table*}

%% file: section/auxlbl.tex
\begin{table*}[h!]
\caption{The ablation study on auxiliary load-balance loss $\mathcal{L}_{\text{aux}}$}
\tabcolsep=5pt
\label{auxloss}
\vskip 0.1in
\begin{center}
\begin{small}
% \begin{sc}

\setlength{\tabcolsep}{4pt}
\resizebox{\columnwidth}{!}{
\begin{tabular}{l|cccccccccc|c}
    \toprule[1.25pt]
    \textbf{Model} & \textbf{ARC-C} & \textbf{BOOIQ} & \textbf{HELLA} & \textbf{LAMBDA} &  \textbf{MNLI} & \textbf{PIQA} & \textbf{RACE} & \textbf{SIQA} & \textbf{WINO} & \textbf{WNLI} & \textbf{AVG.} \\

    \midrule
\multicolumn{3}{l}{\emph{925M Activated 290M}}\\
    \midrule
    
 Traditional MoE \texttt{w/o} $\mathcal{L}_{\text{aux}}$ \type{4E/64E}& 28.33 & 46.85 & 45.63 & 33.44 & 34.58 & 68.88 & 30.72 & 38.69 & 52.49 & 50.70 & 44.66  \\
Traditional MoE \texttt{w/} $\mathcal{L}_{\text{aux}}$ \type{4E/64E} & 30.55 & 56.94 & 47.78 & 32.70 & 34.39 & 69.53 & 30.33 & 39.87 & 52.80 & 40.85 & 45.02 \\
        \midrule
  MoNE \texttt{w/o} $\mathcal{L}_{\text{aux}}$ \type{8E/64E }& 28.67 & 52.72 & 44.93 & 32.06 & 34.85 & 69.48 & 30.14 & 39.92 & 53.91 & 42.25 & 44.47   \\
 
  \rowcolor{Gray}  MoNE \texttt{w/} $\mathcal{L}_{\text{aux}}$ \type{8E/64E }& 30.97 & 55.75 & 48.01 & 33.34 & 34.44 & 70.67 & 29.86 & 38.89 & 53.83 & 49.30 & 46.01 \\

    \bottomrule[1.25pt]
\end{tabular}
}
% \end{sc}
\end{small}
\end{center}
\end{table*}